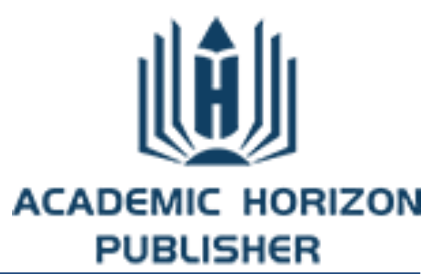

*Perspective*

# A Discussion to Qualify Intelligence

**Kieran Greer** *

Distributed Computing Systems, Belfast, UK
* Correspondence: kgreer@distributedcomputingsystems.co.uk

**Abstract:** Our understanding of intelligence is directed primarily at the human level. This paper attempts to give a more unifying definition that can be applied to the natural world in general and then Artificial Intelligence. The definition would be used more to qualify than quantify it and might help when making judgements on the matter. While correct behaviour is the preferred definition, a metric that is grounded in Kolmogorov's Complexity Theory is suggested, which leads to a measurement about entropy. A version of an accepted AI test is then put forward as the 'acid test' and might be what a free-thinking program would try to achieve. Recent work by the author has been more from a direction of mechanical processes, built from structure. This paper agrees that intelligence is a pro-active event, but also notes a second aspect to it that is in the background and mechanical. The paper suggests looking at intelligence and the conscious as being slightly different, where the conscious is this more mechanical aspect. In fact, a surprising conclusion can be a passive but intelligent brain being invoked by active and less intelligent senses.

## 1. Introduction

Our understanding of intelligence is directed primarily at human beings, but as the meaning of intelligence is undecided, it is difficult to apply the concept to other entities accurately. When writing a paper [1], it became clear that there is no neat or concise definition of what Artificial Intelligence is either. When once asked, I firstly replied with the 'independent behaviour' line, but then added that an 'if-then-else' statement in a computer program might be considered as intelligent. It is able, by itself, to make the decision of what step to take next, even if this is hard-coded. That answer is lacking however and not very helpful. This paper will therefore propose a different definition that would be more helpful for making judgements. In fact, more than one definition is suggested and the discussion is quite open, but certain aspects of intelligence are still quantified. Another goal is to provide unifying definitions, to include the natural world and then the non-biological or artificial life, to include AI. Humans are simply placed at the top of the system when considering these mechanical aspects. To illustrate the lack of a concise description, a few definitions of intelligence from dictionaries are as follows:

*The ability to gain and apply knowledge and skills. (Pocket Oxford Dictionary)*

*(1) The ability to learn or understand or to deal with new or trying situations. (2) The ability to apply knowledge to manipulate one's environment or to think abstractly as measured by objective criteria (as tests). (Merriam Webster)*

*The ability to understand and think about things, and to gain and use knowledge. (macmillandictionary.com)*

*There are probably as many definitions of intelligence as there are experts who study it. Simply put, however, intelligence is the ability to learn about, learn from, understand, and interact with one's environment. (giftedkids.about.com)*

*(1) Intelligence is what you do when you don't know what to do. (2) Intelligence is a hypothetical idea which we have defined as being reflected by certain types of behaviour. (brainmetrix.com)*

Key elements therefore include the ideas of learning, reasoning, understanding and application, to use what is learnt. It is useful to this paper that the concepts of behaviour and environment are also included. What we learn or decide upon should eventually result in some sort of physical event. Therefore, it could be argued that the acts of learning, reasoning and application in fact result in the act of correct behaviour. This is an attractive way of looking at intelligence, because other definitions are based more on human thought, whereas 'behaviour' is a much more gaugeable concept. This is not intended to be Behaviourism, however. One question may then be: why try to define intelligence through behaviour, when intelligent thought is always the problem? It depends on how you want to look at it and what you want to understand about it. For one thing, what is the point of intelligence if you do not try to do something with it. Therefore, it should be reflected in some type of action, but which can include reactive and instinctive processes. For example, human behaviour requires stimulus, creativity, rewards and so on, but maybe some act does not require thinking in the present, but evolved from thinking in the past. It is also the case that different paths can lead from A to B and there is no algorithm to state which way is best. This is why focusing only on the thinking process may be a mistake. Other authors have taken the same approach, but thinking-only should still be accommodated, when the behaviour may become internal, for example, to satisfy a concern. While we may therefore have difficulty measuring intelligent thought, we can measure behaviour more easily, especially when we apply it to other animals. Depending on the level of inspection, internal thoughts can always be considered.

The problems for a concise definition for Artificial Intelligence are clear when considering the arguments from the logicians, such as McCarthy [2]. He used a thermostat argument in a similar way to the if-then-else statement and argued that a thermostat therefore displays intelligence. But he had to point out a whole myriad of technologies and theories that collectively, have to be called intelligence. It seems that trying to define AI in terms of logic, formal languages or processes, shows that this is not what it is. There are too many facets required in a formal description of that type and it is the encompassing entity of all of these different facets that is the AI. Therefore, the definition needs to be extracted to something much more general and simplistic. Another question might be, if we do come up with a concise definition for AI, will that be helpful in a practical sense? If we state that AI means to be 'smart' for example, then can that help to build the system? A practical definition for AI should at least try to guide the development process and act as some sort of a measure for success. Another challenge therefore is to find a definition that can also be used as some sort of metric. The paper will also suggest that intelligence and the conscious are not exactly the same, where the latter term will not be defined to the same level of detail, but will still be given an interesting meaning.

The rest of the paper is structured as follows: Section 2 proposes a definition for a universal idea of intelligence and Section 3 gives some related work. Section 4 considers the mechanical aspects and Section 5 tries to qualify the problem. Section 6 applies the new definition specifically to human-level intelligence, while Section 7 gives some conclusions on the work.

## 2. Levels of Intelligence

All the entities in the Universe are made from the same fundamental materials—natural, man-made, living or not living, etc. This is the domain of the Physicists [3, 4] and as far as intelligence is concerned, suggestions can be made for unifying theories. Because we are all made from the same ingredients, if there is no magic part to intelligence, then it must also be created from those ingredients. It might therefore be possible to apply the concept to all naturally occurring objects. This paper will suggest such a definition, assumed to be for the biological world, and then extend it to man-made ones. If intelligence is defined simply as what a human being would do, then there is a large gap below this that can be comfortably filled with intelligent acts which will not be recognised as such. Looking at the animal kingdom, we can recognise intelligence in other animals through their correct behaviour. We know how they should typically act and are therefore able to notice if they do something wrong. Usually, the benchmark to determine incorrect behaviour is still what we might do ourselves, but that is more a question of 'how'intelligent and not if there is 'any'intelligence, or even the type of intelligence. With human beings we set a higher standard that we can also better understand.

Looking at natural entities that are not part of the animal kingdom is slightly different. Plants, for example, are considered intelligent by some people, but would generally be considered to follow a pre-programmed set of actions, having a nervous system but not a brain. This is also the case for the lower level of the animal kingdom, for the insects possibly, although it is interesting that they can still appear to perform intelligent acts collectively. For this paper, hard-coded intelligence is considered OK and simply at a lower level. This passive type of AI in living entities will also be considered to be more like the conscious. What about something like a rock, sitting on the ground? Is a rock intelligent if it behaves as a rock should? If it does in fact just sit on the ground and slowly decay, then it is doing what is expected. The rock is maybe considered to have a conscious more than an intelligence, where the conscious would be associated with its interaction with the environment. If we throw the rock into a lake, it should sink, but what if it decided to float instead? That incorrect behaviour would be deemed unintelligent – for the rock. You could argue some vague set of concepts, such as the rock knows that water is less dense and the lake bed is dense enough. It knows it is now on water and therefore decides to sink. A new type of rock would need to be recognised, before floating was considered OK and therefore intelligent (pumice). It might be the case that a rock would have very small levels of intelligence and unintelligence, whereas human beings would have much larger levels. Then to re-evaluate the case for something that has changed; doing more than expected as well as less, needs to be considered. Is it an improvement or a mistake? Another condition for intelligence would have to be intention or deliberate acts, which is why the act needs to manifest the thought. The theory starts with 2 definitions for intelligence, as follows:

**Proposition 1.** *An entity can be said to have intelligence if it behaves correctly, inside of the model for which it is defined.* **Proposition 2.** *Artificial Intelligence is then to create this outside of nature, or artificially.*

With behaviour being a measurable quantity and the model being provided naturally. The phrase 'for which it is defined' is very important and assumes some understanding of the constituent parts. Past, present and future events all influence the behaviour and so correctness is not a singular concept and could vary. Some propositions in later sections may help to qualify it further. Because the intelligence can be put into context, any type of natural entity can be considered. While genetics would obviously be an influence, environment is also a key factor. So there is always this juxtaposition of the free-thinking intelligence, backed-up by the mechanical conscious.

With these two propositions, we can start to look at man-made entities, such as computer programs, or hardware that might exist in robots. For example, the propositions state that an 'if-then-else' statement is intelligent if it behaves exactly as that. The statement can make a decision to perform act 1 or act 2, depending on its input. While this decision is hard-coded, a decision is still made and the statement should always perform it correctly. Computer memory devices also appear to be performing complex and useful acts. They store lots of interesting information and are able to retrieve it upon request. However, this is again as far as the model goes. The act of copying and repeating is not a valid description of human intelligence but may be for some computer parts. As an artificial example, robots can be partially intelligent as a whole, but each individual part may be wholly intelligent by itself. This separation is recognised because of the single mind or conscious that we have that is missing in robots. But the conscious is still a distributed system, linked to a centralised and controlling mind. As the single mind understands each distributed part, it is not entirely separate from them.

### 2.1 *What is Not Intelligent then?*

If intelligence is an entity behaving as it should, then 'not' intelligent must be not to behave as the model is defined. If the definition relates to the act specifically, then it is relatively easy to define acts that are not intelligent, such as putting your hand into the fire. The idea of just thinking without acting is more problematic. You may imagine putting your hand into the fire, so long as you do not actually do it. This is considered again in Section 4. While the word intelligent would not often be used for an individual ant, if it did behave differently, we would more easily use a word like stupid. If, for example, it started to move the eggs outside the nest. While this might come naturally from our own superiority, the counter phrase is still often used and so we already associate some form of (non)intelligence with other entities. This is probably the mental process of ascription [5, 6].

## 3. Related Work

This section notes a few papers and arguments, where a complete review would surely include many more.

### 3.1 *Mathematics and Universal Theories*

The term behaviour might be replaced by terms such as common sense [2] or rationality [7–10]. The author thinks that there is a difference however, which is the focus on deliberate acts versus more automatic ones, although, McCarthy [2] notes that innate knowledge, maybe learned through evolution, is part of common sense. The same paper notes the following: 'Shannon's quantitative information theory seems to have little application to the common sense informatic situation. Neither does the Chaitin-Kolmogorov-Solomonoff computational theory. Neither theory concerns what common sense information is.' So, if a distinction can be made between deliberate and automatic, both types of intelligence can be included. While Shannon's paper on Information Theory [11] is mostly a coding problem, the start of the section on 'choice, uncertainty and entropy' does suggest an intelligent system that is trying to understand the information:

> 'We have represented a discrete information source as a Markoff process. Can we define a quantity which will measure, in some sense, how much information is "produced" by such a process, or better, at what rate information is produced? Suppose we have a set of possible events whose probabilities of occurrence are $p_1, p_2, \ldots, p_n$. These probabilities are known but that is all we know concerning which event will occur. Can we find a measure of how much "choice" is involved in the selection of the event or of how uncertain we are of the outcome?'

> 'Theorem 2: The only $H$ satisfying the three above assumptions is of the form:

$$H = -K \sum_{i=1}^{n} p_i \log p_i \tag{1}$$

> where $K$ is a positive constant.'

Higher-level functions representing higher-level algorithms, probably do not have to be continuous and Shannon's information theory assumes a closed event space with a fixed probability distribution, which is not the case for real intelligence. Note however the importance of choice over the information and how that increases the level of uncertainty. Combining information is also preferred to keeping it separate. If neural structures are quite orthogonal, then intelligence creates a collective from it that is more akin to binary associations than weighted combinations. For a sequence of events, $H$ is approximated to be the logarithm of the (reciprocal) probability of a typical long sequence divided by the number of symbols in the sequence. The paper also states: 'If there are no statistical influences extending over more than $N$ symbols, that is if the conditional probability of the next symbol knowing the preceding $(N-1)$ is not changed by a knowledge of any before that, then $F_N = H$. $F_N$ of course is the conditional entropy of the next symbol when the $(N-1)$ preceding ones are known, while $G_N$ is the entropy per symbol of blocks of $N$ symbols.' The paper then states an entropy measure that includes the in-built structure:

> 'The ratio of the entropy of a source to the maximum value it could have while still restricted to the same symbols will be called its relative entropy. This is the maximum compression possible when we encode into the same alphabet. One minus the relative entropy is the redundancy. The redundancy of ordinary English, not considering statistical structure over greater distances than about eight letters, is roughly 50%. This means that when we write English half of what we write is determined by the structure of the language and half is chosen freely.'

'Structure' is the mechanical aspect, whereas 'free' is the intelligent aspect. Kolmogorov extended Shannon's theory and produced a theory that links minimum distance with object complexity. As described in [12] chapter 7, Kolmogorov defined the algorithmic (descriptive) complexity of an object to be the length of the shortest binary computer program that describes the object. Thus, the Kolmogorov complexity of an object dispenses with the probability distribution of Shannon, in favour of a shortest distance. Kolmogorov made the crucial observation that the definition of complexity is essentially computer independent and the shortest binary computer description of a random variable can describe its entropy, or amount of uncertainty. Thus, the shortest computer description acts as a universal code which is uniformly good for all probability distributions. In this sense, algorithmic complexity is a conceptual precursor to entropy, where less complicated algorithms represent the best entropy. This idea is used in Section 5 to try and link entropy with intelligence. The authors then state: 'One does not use the shortest computer program in practice because it may take infinitely long to find such a minimal program.' The chapter then describes that, based on the Turing Machine: 'This led Church to state what is now known as Church's thesis, which states that all (sufficiently complex) computational models are equivalent in the sense that they can compute the same family of functions. The class of functions they can compute agrees with our intuitive notion of effectively computable functions, that is, functions for which there is a finite prescription or program that will lead in a finite number of mechanically specified computational steps to the desired computational result.' This is helpful, because it does not state what those steps are, only that sufficiently complex models can obtain the same result. The theory also states that if a shorter or a longer program produce the same length output, the shorter program is likely to produce the better structure. A smaller change is better and this is known as Occam's Razor, which can be used to define a universal probability theory, which states that in nature, simpler things are more likely than complicated ones and if we wish to describe something, we might consider the simplest description to be the most likely.

The papers [9, 13] go further in trying to quantify intelligence and even propose an equation for a general measure of it. They describe that artificial intelligence can in fact be defined mathematically, in terms of the maths rejected earlier, which is: sequential decision theory and universal induction, which requires Occam's Razor, Kolmogorov Complexity, Turing Machines, Probability Theory (Bayes [14]) and Solomonoff Induction. These are all described in [9] that also gives the full formula. While those measurements can be more exact, the equations later in this paper could be used to try to qualify relative amounts instead. It is interesting however that some of the definitions in [13] are based strongly on the environment and adapting to it (good behaviour). The paper [6] also tries to provide a universal test for intelligence. It overlaps with these theories and includes passive and active environments, but is again much more mathematical. One key point from the paper is that 'intelligence is defined as an average that converges in the limit.' They suggest however that intelligence recognised through adaptive behaviours must be measured through cognitive abilities and not physical acts. But the arguments for a universal test based on adaptive behaviour and even intelligence in simpler forms of life, including plants and even machines, is interesting. They take the view that some form of hard-wired evolution to 'exhibit' a complex behaviour (insect swarms or communities, for example) is false and a truly intelligent behaviour requires individual, non-hardwired adaptation. They also recognise a distinction between mind or consciousness and intelligence. The paper [15] gives a good summary of the problem of defining AI and you could take many quotes from it. For example, Wang writes:

> 'Even a formal definition still needs interpretation when it is applied to a practical situation, and the existence of different interpretations may undermine the exactness of the definition.'

> 'Therefore, the demand for exactness can only be relatively satisfied, as there is no way to completely remove ambiguity in a definition. This is also because some concepts used in the definition may not have exact definitions themselves, and to demand their definitions will cause an infinite regression. No matter how hard we have tried, we have to stop somewhere and depend on some common understanding about some concepts as the starting point of a definition process.'

The definition must address the processes involved, which will eventually lead to evaluating those processes directly. In this respect, the behaviour can be linked with the thought process if it is to be measured. But if it is simply to prove that intelligence exists, then maybe not. But these two facets should probably co-exist. A system should be able to describe itself using Explainable AI [16], for example, when the problem has been formally defined. The system may also be allowed to have a black-box part that simply gets things right or wrong. Black-box functions are a legitimate method when the quantity cannot be easily measured, for example.

### 3.2 *Reactive AI*

Reactive means to react or respond to something, such as your environment. The paper [17] gives a nice description of intelligence: 'Intelligence is the ability to learn from experience and to adapt to, shape, and select environments.' It also summarises other theories that are more on the biological side. It notes that while an 'intelligence gene' has been looked for, no single gene has been conclusively identified. It also notes that race is not a factor. This paper is not interested in creating something by combining both the biological and artificial worlds, such as a cyborg. That is a slightly different domain, where something like [18] might be an introduction to it. Chapters 7 and 8 of that book can also be found in a paper that gives a materialistic view that does overlap with this paper. This paper proposes to separate the concepts of intelligence and consciousness, thereby allowing the natural world to have a conscious, but also allowing the artificial to exhibit high levels of intelligence. The conscious is more in the background, while intelligence is more pro-active. Whole books on consciousness have been written, where one or two references might be [19, 20]. Those chapters quote Damasio when explaining the relationship between emotions and more advanced centres in the brain. 'His message, in brief, is that emotions are both primitive in the sense that we carry around the emotional systems that evolution installed in our brains long before we had warm blood, and that they play intimate roles in all of the higher-level decisions that we tend to think of as rational and emotionless.' The living model therefore uses emotions as part of intelligence, or to 'feed' the intelligence. Damasio's theory [21] notes a three-layered conscious of: emotions, feelings and feeling a feeling. Damasio's definition of emotion is that of an unconscious reaction to any internal or external stimulus which activates neural patterns in the brain. 'Feeling' emerges as a still unconscious state which simply senses the changes affecting the Protoself due to the emotional state. These patterns develop into mental images, which then float into the organism's awareness. Put simply, consciousness is the feeling of knowing a feeling (https://en.wikipedia.org/wiki/Damasio%27s_theory_of_consciousness). Damasio's theory also supports the possibility of a separate conscious and intelligence. The Protoself is the first stage that is a collection of neural patterns that represent the internal body state. They detect and record internal changes that drive the conscious. Section 4.1 describes a scenario where the different body parts provide active input to the brain and compete for resources. The whole-body conscious must balance all of these needs. Wang [10] also uses resources as part of his argument. While Occam or Kolmogorov would promote a minimal solution, it is only minimal in terms of energy use, equating to shorter distances in the brain. Church states that sufficiently complex sequences can tend to give the same result and so optimisation may be at a more local level that is then concatenated. This also puts more emphasis on achieving the final goal, in longer sequences. Even if states merge in the brain, there still needs to be transitions between them for reasoning, and so on, and so some processes require algorithms. How these may be generated is not clear, but naturally separating neural patterns would be a good start and so an orthogonal solution that makes use of binary results should be helpful.

The self-abilities (for example, self-organise, self-heal) are part of autonomic systems [22] and might also set intelligent beings apart from non-intelligent ones. If intelligence is more automatic than we think, then a reactionary element would be important and is probably linked with evolution. While a rock, for example, tends to chaos; self-abilities can allow other organisms to try to mend things as well. Entropy helps to describe this process [23], as the system moves from a state of order to disorder. The author uses some of his own ideas later on, where the paper [24] defined a metric for modelling and measuring autonomous behaviours, which was based on the stigmergic principles of insects, such as termites or ants [25] and included components for both individual and collective capabilities. A similar type of metric is suggested in Section, as a measure of relative intelligence. The authors of [5. 6] note with their equation, that a universal measurement is semi-computable at best, because of the arbitrary way it would be measured. Some of the more classic examples are discussed in the next section.

### 3.3 *Classical AI*

The argument of Section 6, with the idea of developing beyond rote learning, is the sort of argument initially given by Searle [26]. In his 'Chinese room'example: A person with no knowledge of Chinese, can give replies to Chinese questions, by associating symbols that he/she does understand with the Chinese ones and then also using supplied sets of rules to manipulate them. He also noted intention and causal elements (neurons, synapses, nervous system, etc.) as key in human intelligence and the difficulty of creating these artificially. These elements were missing from the computer program that mimicked how the human would try to answer the Chinese room questions. The argument was that following the proposed process would not result in a human 'understanding' Chinese, even if Chinese could be written. Any symbol not included in a rule (not already defined) would not be understood. Therefore, a computer program using the same process cannot learn anything outside of it either, or outside of its programming. The argument however is restricted to early ideas, often based on formal languages and logic, and there have been many developments since then. For example, a real human might be taught exactly what each symbol is first, as a basis for making comparisons and computers have come a long way to understanding natural language. But in practice, algorithmic solutions have not been achieved completely. Therefore, some element of the solution must be missing, which might be some creative aspect. If we learn from our senses, however, then it is assumed that any input is learned and so there is a clear synthesis between the senses and the intelligence here.

Therefore, the manner or way in which the system is taught is also critical. This also relates to the algorithm that would be used to teach the computer [27] and there are differences between a static set of rules and a dynamic system that can change. Searle also quotes McCarthy [5] as stating: 'Machines as simple as thermostats can be said to have beliefs, and having beliefs seems to be a characteristic of most machines capable of problem-solving performance.' He also gives an example of an automatic door with sensors, but appears to be against the idea of intelligence outside of the human mind, or at least outside of a model based almost exactly on it. The belief being that the causal and intentional states of the human brain cannot be duplicated in a computer program. It is not enough to create new knowledge from your existing programming, you need to actually change or expand your existing program.

In his book, Penrose [4] also argues that human consciousness cannot be algorithmic. As that is what computers use and try to copy the world with, the argument then, is that computers cannot be genuinely intelligent. Penrose hypothesizes that quantum mechanics plays an essential role in the understanding of human consciousness. The collapse of the quantum
wave function is seen as playing an important role in brain function (https://en.wikipedia.org/wiki/The_Emperor%27s_New_Mind). This is the phenomenon in which a wave function – initially in a superposition of several eigenstates – appears to reduce to a single eigenstate (by 'observation'). Consciousness is therefore again the combination of several states, factors, or whatever, into a more single whole. If quantum effects cannot be stored in memories however, then the conscious would only happen when the brain is active. Other people have argued that you can simply copy, or even download the brain onto a computer, in which case it is entirely algorithmic.

The paper by Turing [28] can be noted and it does ask about how mechanical our brain processes are. It also suggests a very basic punishment-reward scheme that might replace emotions in a computer, to help with the learning process. The problem of learning, or the program extending itself is written about more than once, where Turing gives a scenario of some type of chain reaction, caused by small disturbances that activate other ones. We now know that this is of course what happens and inhibitors are required to control the firing process, but if enough small regions fire together, then is that enough to give us our resulting thoughts? He also suggests that other animals do not possess the same levels of coherence for combining small regions, making them more subcritical, or less intelligent. The main argument from these papers has therefore been that human intelligence at least, requires a consciousness, which is the synthesis of a distributed architecture.

### 3.4 *Modern AI*

Modern AI is built on Deep Neural Networks [29], Deep Learning [30] and Generative systems [31], such as Large Language Models (LLMs) [32]. A recent study showed that Deep Neural Networks have a built-in Occam's razor-like inductive bias to select the simplest functions, in over-parameterised networks [33]. The bias and a Kolmogorov connection, gives some relation with this paper's objectives. The progress has been at such a rate that some even argue an existential risk to the human race. However, proponents are quick to point out that the systems are still mostly statistical, even though a new property of emergence has been realised in very large distributed systems (LLMs) that is not statistically predictable. Recently, agent-based systems have made a return as Agentic AI [34]. These are autonomous systems that can proactively plan and make decisions, often cooperating and coordinating to achieve complex goals, with multi-step tasks and with minimal human oversight. Image and pattern recognition are all but mastered, as are closed problems with complete information, such as games like Go or Chess. The argument has moved on to more open reasoning processes with incomplete information. It is thought that the systems need to learn how to think in more abstract terms, including a more abstract representation of the data. This has been integrated into a new image recognition system [35], for example. AI however is finding its way into our daily lives as an aid tool at least and a replacement in some cases. What this tells us is that we can get very far with automated processes and that human intelligence has a large statistical component.

## 4. Mechanical or Magic?

So, while our intelligence appears to be unique, you can also ask the question the other way and wonder why intelligence is so routine. Evolution grows a brain most of the time. This section discusses the problem further.

### 4.1 *About the Human*

As we all mutate and change, the intelligence aspect must be more robust than fine details. A more-clear scenario is something like a neural network that generalises over its input, to compensate for noise. The brain stays the same, even after small changes. One argument can be that the small changes are acting independently of each other and while they may have an effect, they do not disrupt the main bulk of the organism, which stays the same. If evolution does use a controlling process to produce consistent results, then mathematical and mechanical theories have been suggested in the research. Another argument is that the problem is simply too complex to have happened randomly in the first place, but if some entities are designed to a consistent model that can provide some level of order, the rest becomes a smaller problem to be solved. If we take the argument that the model is too complex for purely random events, then there must be some mechanical processes to allow it to happen and they would be more reliable. Simply changing the environment could also be a controlling factor, but not reliable. For example, the adage 'no two people see the world in the same way.' If we see objects differently, then at the end of the day when we add that up and put it into our memories, we are likely to make slightly different judgements and this may help the intelligence to grow in different directions.

The mechanical aspect therefore, supports a whole-body conscious. To manage it, a preferable scenario might be if we are made from competing parts that would simply disagree with each other over resources. A single (selfish) theory for one part would have more difficulty dominating the (sub)conscious, where they must all cooperate and share to survive. Interesting then if the conscious can be passive but also a driving force, even if it is not the intelligent part. The intelligent brain is then mostly reactive, to what it gets fed. When we go to sleep, we think less, for example. During normal operations, the body would send stimuli to the brain, where the brain would respond with a stimulus output. What about the actual thought? That must be an understanding of the brain firing patterns, but can the thought re-produce it? For example, a memory of an image starts with the image being projected onto the eye. It then gets stored as a neural pattern, but when we retrieve it again, is the eye still used? If it is missing, then the neurons need to be able to produce exactly what the retina has. If the eye is used during the feedback, then it can be the interpreter. Without the eye it would be a very magic process to re-produce the image. And when thinking of an image internally, can we see properly at the same time? We
would normally attribute this to thinking about two things at the same time, or sense overload.

### 4.2 *About the Robot*

A future test might measure the level of sentience, through some measure of pain or discomfort, which would be linked to having a conscious. Emotion in machines has not been discarded, but it is generally thought of as something that they generate through an algorithm. An AI creation that is as emotionally fragile as a human is not normally thought about, but do we know how the conscious would work in a machine? With increasing levels of intelligence, would it become emotionally attached to its body parts? Will that level of singular understanding ever exist, maybe if synthesis is involved. If not, then why may it become more intelligent than we are, because if intelligence is separate from the conscious, then computation power is a factor. The current climate is producing lots of scare stories about the dangers of AI, especially in the Military. So one question might be, if we do everything correctly and properly and build the AI robot, will it 'naturally' attempt to take over? If the robot doesn't have the selfish gene or emotional state, then would it wish to, or would it even have a survival instinct? That might be critical. Would it be able to dominate through logic alone, without emotional help?

The assumption is that it will simply know too much and find human requests illogical. While it might realise a comparable level of intelligence, it would probably be from a different type of mechanism. Our own model relies on the unifying states. It would appear to be what makes us superior and as we do not fully understand it, we cannot create it artificially. Instinct is important as well, possibly through a long evolutionary process and that is obviously hard-coded. If the programming environment is too coarse-grained to be successful, maybe it is easier to let the computer evolve itself. So the living model has an ideal combination for creating the higher levels of intelligence and for the artificial model, we need to duplicate that.

## 5. An Intelligence Metric

This section tries to qualify AI by proposing a very basic metric and algorithm that may be related to Information Theory [11] and in particular, the resulting theory by Kolmogorov ([12], chapter 7). This is a measure of whether a system exhibits intelligence, but not to quantify it accurately. The metric would try to recognise relative intelligence amounts,
where for a comparison, the description from Wang [15] seems to be appropriate. A first working definition, proposed in 1995 was:

'Intelligence is the capacity of an information-processing system to adapt to its environment while operating with insufficient knowledge and resources.'

That definition is very similar to this paper. It can include good behaviour in an adaptive system, but the system does not want to forget about earlier environments when adapting. This was then extended into the AIKR definition, known as the 'Assumption of Insufficient Knowledge and Resources,' which identifies the normal working condition of an intelligent system. However, it is based strongly on the problem-solving aspect where: 'with insufficient knowledge and resources' is divided into three restrictions or requirements, that is – the system must be finite and open, and operate in real-time. The problem is when we try to quantify intelligence. Rational and learning are key concepts, but they must be balanced against more passive methods. Rational relates more to intellect than experience or emotion, but all may be needed for the conscious. At this point, it may also be useful to recognise the difference between a good and a bad environment, where Wang writes about circumstances, as well as absorbing the experience into the system. The passive (emotional and experience-based) AI running in the background can be over-ridden by rational and more pro-active thought. Defeasible logic is a weaker type of logic that can be annulled depending on the result, for example. It is likely however that there is more deliberation in the bad environment and intelligence is more automatic when in a good environment. In fact, if everything is good, then there is not much reason to change and a more intelligent system will already know the answer. This leads to another statement that can help to define an intelligent system:

**Proposition 3.** *A more intelligent system will try to stay the same, to reduce entropy and energy, and also to remember what it already knows.*

Then there is the question of how to measure if the system exhibits intelligence. If intelligence is related to the amount of change, then that is a pro-active activity which might alter background processes as well. But the change should be minimal and after a change, the circumstance leading to it should be lessened, meaning that any change for the next step should consequently also be less. Therefore, this argument can be used as a measure, to determine if the system exhibits intelligence, as follows:

$$I_{t+1} = I_t + c_1,$$

$$I_{t+2} = I_{t+1} + c_2,$$

$$I_{t+3} = I_{t+2} + c_3, \quad (2)$$

where $c_3 < c_2 < c_1$.

In this case, the changes $c_3 < c_2 < c_1$ would be the trend to indicate that the system is exhibiting some intelligence, where the result at time $t+1$ is the result at time $t$ plus the change. If that was not happening and the behaviour could be altered, then a search process could try to do that. This is therefore, also a measure of reducing the uncertainty, or problem to be solved. There is also likely to be a horizon effect, where at some point in the solution, a whole new set of problems are identified. But the nature of it being new, means that the metric can then be reset. The behaviour metric of [24] may offer similarities with these equation sets. The 'conscious + intelligence' scenario could even be modelled by something like:

$$\text{Entity Model} = \text{Intelligence}+\text{Emotional}+\text{Experience}+\text{Rational}. \quad (3)$$

The behaviour metric was for basic entities, like insects. Intelligence comprised of ability plus flexibility, whereas the full entity design also required collective skills like cooperation, coordination and communication. In the more advanced human design, the collective skills could map as follows: cooperation is emotional, coordination is rational and communication is experience. This would place rational as part of the conscious, or something with instinct.

## 5.1 *Algorithmic Solution*

Considering only intelligence again, the argument has led back to one about entropy, but before describing the equation, the theories that have led to it can be discussed further. Shannon's Information Theory [11] is for a closed space with a continuous distribution, and while Kolmogorov [12] requires that the algorithmic model or object can be coded in binary form, his conclusions can still be applied to the real world. Shannon's theory has some roots in intelligence, with the goal to reduce entropy through more intelligent choices and Markoff processes. While Kolmogorov's theory is abstract, it helps to define that shorter distances are more reliable and more likely. They are also more economic energy-wise. If we are dealing with smaller entities, then we really require a continuous solution for them to be meaningful. A continuous function also means less change and so built into that is the idea of a shorter distance. As the entity gets more complex and larger however, it is allowed to have parts that may be discontinuous to each other and the transition is also from a closed to an open system. The algorithmic solution to it may then be discontinuous between parts, or modular, but preferably continuous inside of each part. But this is what the more intelligent entity can solve and it is the basis for artificial neural networks [36], for example. The system would then take the results of the smaller parts, as new parts and solve those again, as a continuous new problem.

This balancing act is therefore what the more intelligent system can try to solve. A very basic model can be described for doing this, which is statistical and may again have the aspirations of early neural network modelling [37]. If the two information theories are considered to be parts of the same system, then something analogous to Kolmogorov's algorithm would move a process in steps to some evaluation phase and something analogous to Shannon's information theory would then evaluate that phase. It might be interesting to think that an evaluation phase is likely to be repetitive and closed, where the tasks repeat until a satisfactory conclusion is achieved. This is not the case for the Kolmogorov phase, which simply wants to get from A to B in the shortest number of steps. In fact, if the sequence in the Kolmogorov phase repeats, then there is likely to be an error. The repetition would cycle the process backwards, when it would have to move through those stages again and it would not be optimal. The Kolmogorov phase is therefore continuous in a line rather than a loop. This process is illustrated in Figure 1. The Kolmogorov process moves the algorithm in single steps to an evaluation phase, where the Shannon process evaluates the phase, possibly using a Markoff process instead of the minimum straight line, before the next Kolmogorov stage. The Shannon process may be repetitive and should satisfy the result in Equation (2). It is described in the next section but is a Bayes process, rather than a Likelihood.

### 5.1.1 *Example*

An example of this process could be the following: Jack wants to hammer a nail into a post. To do this, he needs to get a hammer and a nail, walk over to the post and hammer the nail in. Kolmogorov would state that he should go straight to where the hammer and nail is, collect them and then go straight to the post. If he made a detour for no reason, then it would be less intelligent. If he made a detour because he wanted to look at some scenery, maybe for health reasons, then that would get factored into the process and a balance would be made. When at the post, hammering the nail in is repetitive, but each step should move the nail further in, indicating progress and less uncertainty. This is more functional and can be measured as such, with good or bad results.

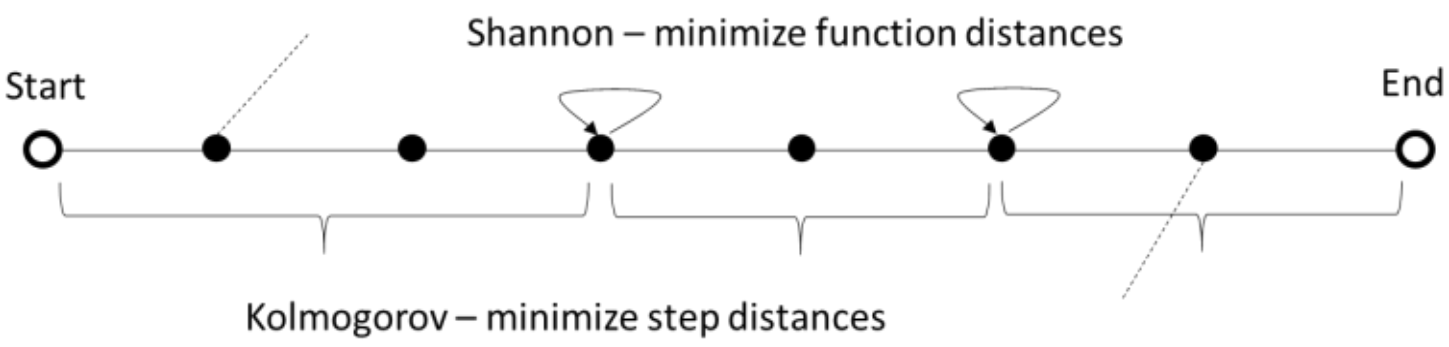

**Figure 1.** Continuous and shortest distances, for the neural path and evaluation.

## 5.2 *Entropy Metric*

Considering entropy therefore, the more intelligent system would take fewer steps, thereby minimising this effect. Also written by Wang [15], the equation should also consider the 'potential' for change, because a system may be very restricted and therefore only have limited capabilities. A system that has many options however should not be penalised if most of them are irrelevant to the current situation and so it is only the relevant set that should apply. In fact, it should be rewarded if it can correctly identify the relevant subset and if that set is useful. An entropy metric can therefore be suggested that would recognise the system intelligence as being proportional to the amount of change that it could apply, divided by a factor of the amount applied and the total amount of possible change, as follows:

$$I_c \propto \left( \sum_{t=1}^{n} \frac{Cp_t}{C_t \cdot Ca_t} \right) \Big/ n, \tag{4}$$

where:

- $I_c$ = Intelligence change
- $C_t$ = change that is applied at time $t$
- $Cp_t$ = potential change that is relevant at time $t$
- $Ca_t$ = total possible change if everything is included at time $t$
- $t$ = time step
- $n$ = number of time steps to make the evaluation over

And a smaller $I_c$ is preferred.

The total possible change and the amount finally applied may be measurable, but the potential or relevant amount might not be as obvious. There are some similarities with Bayes'theorem [14]. Both have 3 evidence quantities, but in this metric, the 'observed' act is a denominator. Also with Bayes, it may be difficult to calculate the conditional probabilities. If the situation is resolved, then it does not really matter, but if not, then another time step and resolution process will be required, where the potential pool to select from may be corrected. Therefore, to compensate the inaccuracy in measuring this quantity, it can be measured over several time steps and averaged, when we have the following proposition based on entropy:

**Proposition 4.** *Intelligence is represented by a minimum amount of change for a new situation, or the minimum entropy change, as measured against the ability to change.*

This in fact looks like an entropy version of Wang's description, but with the inclusion of a minimal amount of change. Inherent in this is an intelligent selection process, which is an intelligent decision process.

#### 5.2.1 *Example*

Jack and Jill are solvers, where the problem has a model of A and C. Jack is a trainee and has a skill set of A or B. Jill is an expert and has a skill set ofA, B, C, D or E. In the first step, Jack uses skillAand in the second step he uses skill B. Jill uses skill E in the first step and skill C in the second step. Both were correct in one step and incorrect in one step, so who is behaving more intelligently? Jack has a restricted skill set, where the potential set for step 1 has 1 skill, but 0 skills for step 2. While it is not clearly defined, the total change for Jack should be set to 1, but for Jill it is 2. The first step for Jack gave a positive change but the second gave a negative change. The equation would give a score of $+(1/(1\times1))-(0/(1\times1))$. Jill had more choice with a full potential skill set, but also more uncertainty about what skill to use. The equation for Jill would be $-(1/(1\times2))+(1/(1\times2))$. This would be negative in the first step and positive in the second, resulting in a net score of 0. If the other abilities are equal, then Jack did better to get something right with his skill, but the equation would also flag a null result for the second step. Remember that the measurement is a relative one about progress (Equation (2)), thereby exhibiting intelligence. A lack of progress could mean poor judgement or that new skills are required and so the equation is intended to help with that decision.

## 6. Human-Level Artificial Intelligence

The Introduction gave some definitions for intelligence and Section 2 proposed to base a universal type of intelligence on the behaviour. Further definitions and rules have also been given and this section returns to the problem of a general definition for Human-level Artificial Intelligence. This would typically try to compare a machine's potential with a human. If an aware or conscious program can be written, intelligence will result from correctly evolving the model. Inherent in this is the ability for the program to learn, for if it is to be self-aware over something new, it must be able to understand the new model first. Many films and programs have shown the Artificial Intelligence program overruling its programming, to take control. This could even be considered as the sort of 'acid test' required for a higher level of Intelligence:

**Proposition 5.** *An entity can be said to have a higher level of (artificial) intelligence if it can correctly and consistently overrule its environment.*

The definition by Sternberg [17] (start of Section 3.2) writes about the ability to shape and select one's environment. This could apply to natural or man-made objects alike and overrule does not mean to conquer, but it might mean to preserve. Any overrule would have to be consistent, to remove the possibility of a random act. This would also imply that the program was aware of what it was doing. If we are defined by our model, then it would be more obvious to state that higher intelligence results from being able to break the model's coding, but that is not so clear. Depending on the level of inspection, you can argue that everything we do is still down to our programming. Breaking our model could mean breaking the environment that has created us, thereby changing its effect.

## 7. Conclusions

This paper has attempted to give a definition for intelligence that can put it in a more general context from the idealised human view. Emphasis is probably now more on a physical or internal act, through correct behaviour, rather than the mental thought. In a large sense, this means that the intelligent system is trying to resolve its environment. This could be linked to awareness of self and situation, but has to allow for simply thinking, with an internal result. The discussion then considered how the intelligence might be created through more mechanical means, which can be linked to the model and used for any natural world entity. A non-living entity should still be governed by the same universal laws and even if its model is hard-wired, it may in some sense have a conscious through interaction with the environment. The concept of consciousness spans many different disciplines, from philosophy to engineering, where we have assumed that we must comprehend this aspect before human-like intelligence can be built. For a living being, it may be associated more with mechanical processes that can invoke change. How an AI system could synthesise this is not clear. As well as a conscious, therefore, we have the raw compute of intelligence itself. This is linked to processing information and Shannon noted that in natural language, possibly 50% of what we produce is determined by the structure and 50% is free. Possibly 50% of intelligence is structural.
Kolmogorov and others then offer optimisation methods to help with formulating these intelligence structures.

If we measure intelligence through material results therefore, our thoughts are still critical to that. Does that make the argument any clearer or easier to evaluate? The question of intelligence still exists, but it is now placed in the context of concepts that we can understand and measure, and not the 'black box' that we guess about. We are also making the measurement more materialistic, where in doing so it loses some relevance, even to the point of a win-loss result. The autonomous system has been mentioned and the idea of an intelligent but dumb brain, being managed and controlled by unintelligent but active sensors is interesting. The brain is the autonomous part that reacts to input and it is also the intelligent part. The pro-active sensors are less intelligent, even though they determine what the brain learns and to some extent, what it thinks about. They represent the conscious more. Both parts therefore are required for a complete system. In fact, a mechanical conscious is helpful in some ways, because it can free the brain from this aspect, which allows it to be more free-thinking. Damasio's Protoself appears to use the same philosophy and is a collection of neural patterns that represent the internal body state. The argument in this paper is that these distributed components then compete for resources, which requires a whole-body conscious to balance all of the needs. The brain is obviously functional, but it is also algorithmic. Each algorithmic part may satisfy some constraint or requirement and the optimising methods may act locally on each algorithmic part. The problem is simplified from weighted distributions to binary sequences, but over longer sequences, it is the win-loss result that matters most.

Then a metric has been proposed, along with some other descriptions and it is possible that it could be used for relative comparisons at least. The entropy argument looks convincing. The third proposition of Section 5 implies that self-aware, conscious, or sentient entities can override their programming. For non-sentient entities this would not be possible, which is why a plant always behaves like a plant, or a rock like a rock. An intelligent being requires more than just a conscious. One might ask just how much intelligence is pre-programmed. Are we mainly running on automatic ourselves? We are supposed to have a selfish gene [38], but that is more for self-preservation than selfish acts. If there are conflicting possibilities however, then the selfish nature of a person might be a deciding factor. Can a machine therefore behave selfishly, or break its model? Can it ever behave outside of the model that originally defined it? There are probably different ways to test for an AI program that can behave intelligently. Five propositions have been suggested in this paper. If these are put into a single statement, then a final definition for intelligence might be as follows: **Definition of Intellegence:** *An entity can be said to have intelligence if it behaves correctly, inside of the model for which it is defined and can maintain this in competing environments. It should be aware of both success and failure, being successful with a minimum amount of effort, but being able to adapt to change after failure.*

## Conflict of Interest

There declare no conflict of interests.

## References

[1] Greer K. Turing: then, now and still key. In: *Artificial Intelligence, Evolutionary Computation and Metaheuristics (AIECM) – Turing 2012*. Springer-Verlag Berlin Heidelberg; 2013. p.43–62. Available from: https://doi.org/10.1007/978-3-642-29694-9_3.

[2] McCarthy J. The philosophy of AI and the AI of philosophy. In: *Philosophy of Information*. North-Holland; 2008. p.711–740. Available from: https://doi.org/10.1016/B978-0-444-51828-3.50026-4.

[3] Cox B. *Wonders of Life*. Wonders Series, Harper Design; 2013.

[4] Penrose R. *The Emperor's New Mind*. Oxford: Oxford University Press; 1989. Available from: https://global.oup.com/academic/product/the-emperors-new-mind-9780198519737.

[5] McCarthy J. Ascribing mental qualities to machines. In: *Philosophical Perspectives in Artificial Intelligence*. Humanities Press; 1979.

[6] Dowe DL, Hernandez-Orallo J. How universal can an intelligence test be? *Adaptive Behavior*. 2014; 22(1): 51–69. Available from: https://doi.org/10.1177/1059712313510660.

[7] Simon HA. *Models of Man: Social and Rational*. New York: John Wiley; 1957.

[8] Russell S. Rationality and intelligence. *Artificial Intelligence*. 1997; 94(1–2): 57–77. Available from: https://doi.org/10.1016/S0004-3702(97)00026-X.

[9] Hutter M. *Universal Artificial Intelligence: Sequential Decisions based on Algorithmic Probability*. Berlin: Springer; 2005. Available from: https://doi.org/10.1007/b138233.

[10] Wang P. The assumptions on knowledge and resources in models of rationality. *International Journal of Machine Consciousness*. 2011; 3(1): 193–218. Available from: https://doi.org/10.1142/S1793840811000316.

[11] Shannon CE. A mathematical theory of communication. *The Bell System Technical Journal*. 1948; 27(3): 379–423; 27(4): 623–656. Available from: https://doi.org/10.1002/j.1538-7305.1948.tb01338.x.

[12] Cover TM, Joy AT. *Elements of Information Theory*. John Wiley & Sons, Inc.; 1991. Available from: https://doi.org/10.1002/0471200611.

[13] Legg S, Hutter M. Universal intelligence: a definition of machine intelligence. *Minds and Machines*. 2007; 17(4): 391–444. Available from: https://doi.org/10.1007/s11023-007-9079-x.

[14] Rouder JN, Morey RD. Teaching Bayes' theorem: strength of evidence as predictive accuracy. *The American Statistician*. 2018. Available from: https://doi.org/10.1080/00031305.2018.1518632.

[15] Wang P. On defining artificial intelligence. *Journal of Artificial General Intelligence*. 2019; 10(2): 1–37. Available from: https://doi.org/10.2478/jagi-2019-0002.

[16] ArrietaAB, Díaz-Rodríguez N, Del Ser J, BennetotA,Tabik S, BarbadoA, García S, Gil-López S, Molina D, Benjamins R, Chatila R. Explainable artificial intelligence (XAI): concepts, taxonomies, opportunities and challenges toward responsible AI. *Information Fusion*. 2020; 58: 82–115. Available from: https://doi.org/10.1016/j.inffus.2019.12.012. [17] Sternberg RJ. State of the art. *Dialogues in Clinical Neuroscience*. 2012; 14(1): 37–49. Available from: https: //doi.org/10.31887/DCNS.2012.14.1/rsternberg.

[18] Brooks RA. *Flesh and Machines: How Robots Will Change Us*. Pantheon Books; 2002.

[19] Chalmers DJ. *The Conscious Mind: In Search of a Fundamental Theory*. Oxford University Press; 1997.

[20] DamasioAR. *The Feeling of What Happens: Body and Emotion in the Making of Consciousness*. New York: Harcourt Brace Jovanovich; 1999.

[21] Bosse T, Jonker CM, Treur J. Formalisation of Damasio's theory of emotion, feeling and core consciousness. *Consciousness and Cognition*. 2008; 17(1): 94–113. Available from: https://doi.org/10.1016/j.concog.2007.04.001.

[22] IBM. An architectural blueprint for autonomic computing. IBM and Autonomic Computing; 2003. Available from: https://www.research.ibm.com/autonomic/manifesto/autonomic_computing.pdf.

[23] Feynman RP. *The Feynman Lectures on Physics*. Vol. 2 (2nd ed.). Addison-Wesley; 2005. Available from: https: //doi.org/10.1016/B978-0-08-050703-8.50001-6.

[24] Greer K. A metric for modelling and measuring complex behavioural systems. *IOSR Journal of Engineering (IOSRJEN)*. 2013; 3(11): 19–28. Available from: https://doi.org/10.9790/3021-03111928.

[25] Garnier S, Gautrais J, Theraulaz G. The biological principles of swarm intelligence. *Swarm Intelligence*. 2007; 1(1): 3–31. Available from: https://doi.org/10.1007/s11721-007-0004-y.

[26] Searle JR. Minds, brains, and programs. *Behavioral and Brain Sciences*. 1980; 3(3): 417–457. Available from: https://doi.org/10.1017/S0140525X00005756.

[27] HoffmannA. Can machines think? an old question reformulated. *Minds & Machines*. 2010; 20(2): 203–212.Available from: https://doi.org/10.1007/s11023-010-9193-z.

[28] Turing A. Computing machinery and intelligence. *Mind*. 1950; 59(236): 433–460. Available from: https://doi.org/10.1093/mind/LIX.236.433.

[29] Hinton GE, Osindero S, Teh YW. A fast learning algorithm for deep belief nets. *Neural Computation*. 2006; 18(7): 1527–1554. Available from: https://doi.org/10.1162/neco.2006.18.7.1527.

[30] Mnih V, Kavukcuoglu K, Silver D, Rusu AA, Veness J, Bellemare MG, Graves A, Riedmiller M, Fidjeland AK, Ostrovski G, Petersen S, Beattie C, Sadik A, Antonoglou I, King H, Kumaran D, Wierstra D, Legg S, Hassabis D. Human-level control through deep reinforcement learning. *Nature*. 2015; 518: 529–533. Available from: https: //doi.org/10.1038/nature14236.

[31] Sengar SS, Hasan AB, Kumar S, Carroll F. Generative artificial intelligence: a systematic review and applications. *Multimedia Tools and Applications*. 2025; 84(21): 23661–23700. Available from: https://doi.org/10.1007/ s11042-024-19602-6.

[32] Minaee S, Mikolov T, Nikzad N, Chenaghlu M, Socher R, Amatriain X, Gao J. Large language models: a survey. *arXiv preprint arXiv:2402.06196*. 2024. Available from: https://arxiv.org/abs/2402.06196.

[33] Mingard C, Rees H, Valle-Pérez G, Louis AA. Deep neural networks have an inbuilt Occam's razor. *Nature Communications*. 2025; 16(1): 220. Available from: https://doi.org/10.1038/s41467-025-55745-5.

[34] Bandi A, Kongari B, Naguru R, Pasnoor S, Vilipala SV. The rise of agentic AI: a review of definitions, frameworks, architectures, applications, evaluation metrics, and challenges. *Future Internet*. 2025; 17(9): 404. Available from: https://doi.org/10.3390/fi17090404.

[35] Assran M, Duval Q, Misra I, Bojanowski P, Vincent P, Rabbat M, LeCun Y, Ballas N. Self-supervised learning from images with a joint-embedding predictive architecture. In: *Proceedings of the IEEE/CVF Conference on Computer Vision and Pattern Recognition*. IEEE; 2023. p.15619–15629. Available from: https://doi.org/10.1109/CVPR52729.2023.01563.

[36] Rojas R. *Neural Networks: A Systematic Introduction*. Springer-Verlag, Berlin; 1996.

[37] Amari SI. Mathematical foundations of neurocomputing. *Proceedings of the IEEE*. 1990; 78(9): 1443–1463.Available from: https://doi.org/10.1109/5.58307.

[38] Dawkins R. *The Selfish Gene*. New York City: Oxford University Press; 1976. Available from: https://doi.org/10.1093/oso/9780198575191.001.0001.